\documentclass[journal,twoside,web]{ieeecolor}
\usepackage{generic}
\usepackage{cite}
\usepackage{amsmath,amssymb,amsfonts}
\usepackage{algorithmic}
\usepackage{graphicx}
\usepackage{algorithm,algorithmic}
\usepackage{textcomp}
\usepackage{subcaption}  
\usepackage{adjustbox}   
\usepackage{booktabs}    
\usepackage{hyperref}
\hypersetup{hidelinks}
\usepackage[capitalize]{cleveref}

\def\BibTeX{{\rm B\kern-.05em{\sc i\kern-.025em b}\kern-.08em
    T\kern-.1667em\lower.7ex\hbox{E}\kern-.125emX}}
\begin{document}

\title{Graph Mamba Survival Analysis Based on Topology-Aware ordering}

\author{
Yuanfang Chen\textsuperscript{a},
Peiqiang Yan\textsuperscript{a},
Yuntao Shou\textsuperscript{b},
Qian Zhao\textsuperscript{a},
Xiangyong Cao\textsuperscript{b}
\\[1em] 
\small 
\textsuperscript{a}School of Mathematics and Statistics, Xi'an Jiaotong University, West China Science and Technology Innovation Harbor, Xi'an, 710049, Shaanxi Province, China\\
\textsuperscript{b}School of Computer Science and Technology, Xi'an Jiaotong University, West China Science and Technology Innovation Harbor, Xi'an, 710049, Shaanxi Province, China
}






\maketitle

\begin{abstract}
    In computational pathology, Whole Slide Images (WSIs) survival analysis is crucial for patient prognosis assessment, but it faces multiple technical challenges. Although the Transformer captures long-range dependencies through its self-attention mechanism, its $O(N^2)$ time complexity causes a severe computational bottleneck in large-scale WSIs graph structures. 
    The Mamba model breaks through the Transformer's computational bottleneck with linear complexity. But, owing to Mamba's high sensitivity to the order of input data, traditional node sorting methods in Graph Mamba, such as those based on node degree or subgraph size, fail to adequately account for the topological connectivity of graph data. This inadequacy consequently restricts the performance of Mamba's sequential modeling. Moreover, its unidirectional architecture cannot leverage the bidirectional spatial structure of images.
    To address these challenges, this paper proposes a novel Graph Mamba survival analysis framework based on topology-aware ordering (TopoMamSurv) to adapt to the sequential sensitivity of Mamba. Our visualization experiments further confirmed that the nodes extracted through the topology-aware ordering (TAO) strategy indeed exhibit higher similarity. Furthermore, we designed a bidirectional Mamba module and integrated a Graph Convolutional Network (GCN) to achieve bidirectional spatial context modeling of images, forming a hierarchical feature learning architecture for "local aggregation - global capture." This framework effectively reconciles the contradiction between long-range dependency modeling, computational efficiency, and spatial structure utilization in WSIs analysis through its systematic design of TAO, bidirectional semantic modeling, and hierarchical feature fusion. This framework has been validated for its comprehensive performance advantage on five TCGA datasets.
\end{abstract}


\section{Introduction}
Survival prediction is a method used to assess patient prognosis by quantifying the time interval from disease diagnosis or treatment initiation to the occurrence of specific clinical events, such as death or disease recurrence. It plays a crucial role in clinical oncology and medical practice by enabling accurate risk stratification and prediction, which assists physicians in developing personalized treatment plans and supports patients in making informed decisions based on prognostic information. Given that cancer remains one of the leading causes of death worldwide, the development of precise and robust survival prediction methods is of great importance for improving clinical outcomes.

With the development of digital pathology technology, Whole Slide Images (WSIs) have become a crucial data source and are of particular interest for survival analysis, thanks to their high resolution and comprehensive tissue morphological information~\cite{lv2022transsurv,liu2023graphlsurv,shao2023hvtsurv}. In computational pathology, due to the large size of WSIs, they are often divided into smaller patches for further processing. Consequently, the key is how to aggregate the patch information to achieve WSI-level prediction~\cite{shao2021transmil}.

Currently, there are mainly two kinds of methods for WSI-based survival analysis, i.e, the multiple instance learning (MIL) methods and the graph-based ones. Multiple Instance Learning (MIL)~\cite{ilse2018attention,li2021dual,xie2020beyond,guan2022node} treats the input patches of a WSI as the instances sharing the same label of this WSI, and uses aggregated patch-level information for patient-level prediction. As a comparison, graph-based methods treat each WSI as a graph with the patches as its nodes~\cite{chen2021whole,li2018graph,liu2023graphlsurv}, and adopt the Graph Neural Networks (GNNs) to obtain the patient-level features for prediction. Since both the local similarity between patches and the global structure of the WSI can be well characterized by the graph, a good representation is expected to be learned, thereby leading to accurate survival prediction among the state-of-the-art ones~\cite{wang2024dual,li2018graph}.

\begin{figure*}[t!]
  \centering
  
    \includegraphics[width=\textwidth, height=10cm]{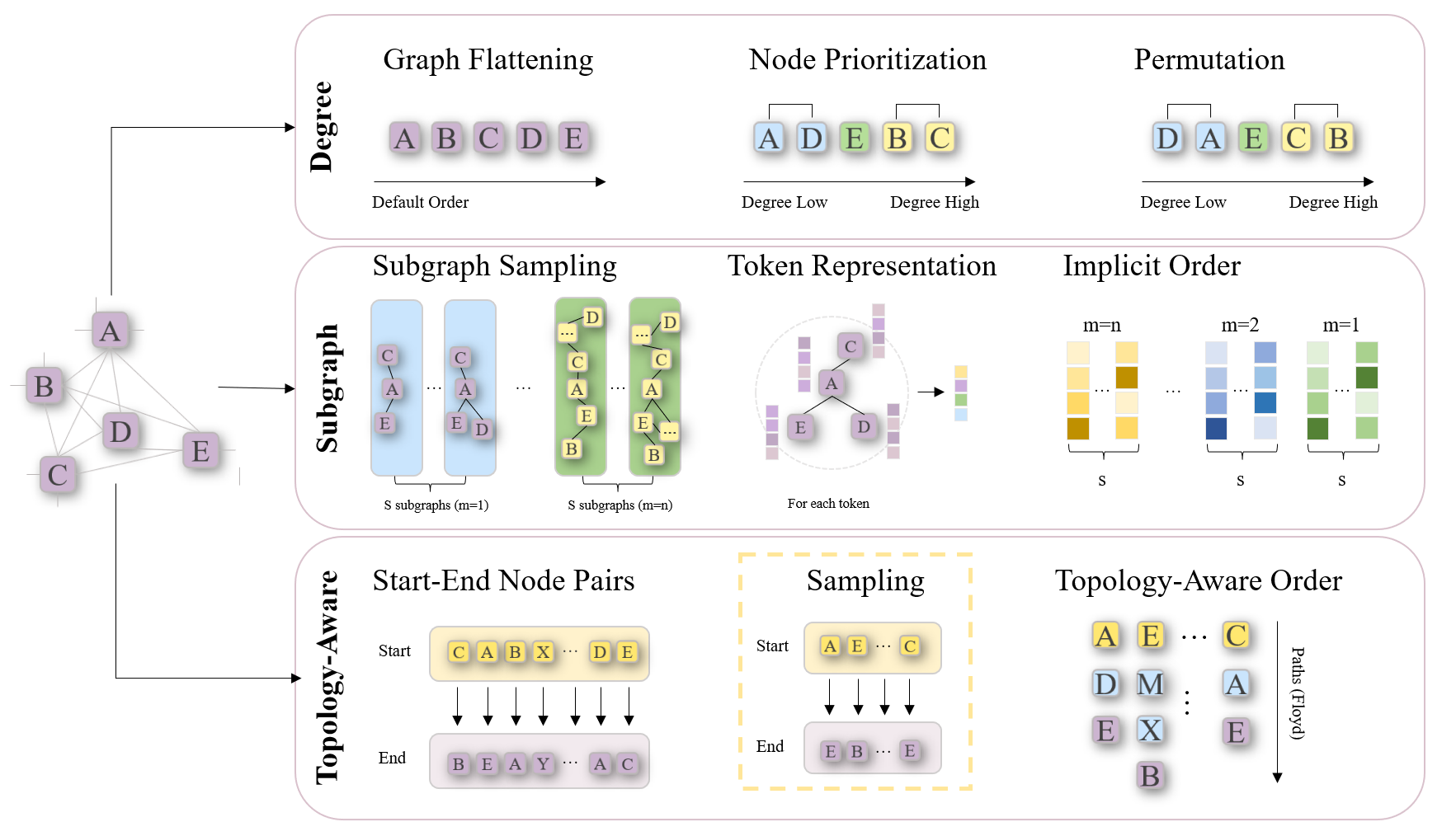} 

  \caption{\textbf{Three ordering methods in Graph Mamba.} 
  \textbf{Degree ordering.} 
  First, flatten the graph; then, sort nodes by degree ascending, and randomize ties.
  \textbf{Subgraph ordering.}
  First, sample \textbf{s} subgraphs via random walk, represent each by a single feature, then sort them in descending order of walk length m.
  \textbf{Topology-Aware ordering.} 
  Enumerate all source–target pairs, optionally subsample for speed, then compute shortest paths with Floyd and list them in order.
  }
  \label{fig:Three order}
\end{figure*}

Though having achieved promising results, GNNs still have limitations. For example, a major drawback of GNN is that the message-passing mechanism is limited to local node neighborhood information aggregation, making it difficult to capture long-range dependencies globally~\cite{zheng2022graph}. To address the issue of capturing long-range dependencies, some studies have attempted to combine the global reception field of Transformers with the message passing of GNN, to simultaneously capture local and global relationships~\cite{liu2022pyraformer}. However, this approach exacerbates the high computational complexity problem due to the quadratic computational complexity of the self-attention mechanism, especially in large-scale image applications like computational pathology~\cite{wang2024graph}. In order to capture global long-range dependencies while reducing computational complexity, the Mamba model has been introduced to computational pathology modeling for large-scale graphs~\cite{yang2024graphmamba}. However, since Mamba is originally a unidirectional sequence-to-sequence model \cite{gu2023mamba}, multidimensional data, such as images or point clouds, should be flattened into a one-dimensional sequence when processing, and thus, the ordering of the sequence for scanning could significantly affect the final performance due to the selective structural memory mechanism. When applying to graph data, such as WSIs, the scanning order becomes more essential due to the complex topological structure of graphs. Currently, several ordering strategies have been adopted for graph-based Mamba, including sorting by node degree \cite{wang2024graph, behrouz2024graph} or subgraph size \cite{behrouz2024graph}. However, none of them considers the topological connectivity of the graph, and thus limits the overall performance.

To address the aforementioned issue of applying the graph Mamba-based methods to survival analysis, we propose a topology-aware ordering (TAO) strategy for graph Mamba based on the shortest-path . Specifically, we propose to construct the scanning sequence by searching for the shortest path between two nodes within the graph. Such a strategy can, on the one hand, ensure that the neighboring nodes along the sequence are topologically true neighbors in the graph, retaining the local similarity structures of the graph; and on the other hand, capture the long-range correlations between the two end nodes. With the proposed scanning ordering strategy, we construct a multi-branch graph Mamba-based architecture that also integrates the Graph Convolutional Network (GCN) structure and the bidirectional Mamba techniques, for WSI-based survival analysis. To summarize, the main contributions of this paper are as follows:

    \begin{itemize}
        \item We propose a topology-aware ordering strategy for graph Mamba based on the shortest-path ordering. Compared with existing scanning ordering strategies, the proposed one can result in scanning sequences that better capture both the local and long-range correlations of the graphs.        
        
        \item Applying the proposed scanning strategy, we present a multi-branch bidirectional graph Mamba-based architecture, with the aid of GCN, for the WSI-based survival analysis task.        

        \item We apply our model to real WSI-based survival analysis tasks, and compare it with existing state-of-the-art methods, demonstrating its effectiveness. We also specifically verify the reasonability of the proposed scanning strategy.          

    \end{itemize}

\section{Related Work}
\subsection{WSI-based Survival Analysis}\label{sec:WSI_SA}
Computational Pathology is a discipline that utilizes artificial intelligence techniques to analyze histopathological images, with survival analysis being one of its important branches. Survival analysis aims to predict patient survival times and provide evidence for clinical decision-making. Traditional survival analysis methods rely on pathologists' visual inspections and manual feature quantification, which are time-consuming and highly subjective. With the development of deep learning technologies, the WSI-based survival analysis has become a popular research direction. 

Deep learning methods can automatically learn complex feature representations from histopathological images, thereby improving the accuracy and robustness of survival analysis. Currently, there are mainly two ways to deal with the WSI-based survival analysis task, i.e., MIL methods and graph-based methods, and we briefly review them in the following.

\begin{figure*}[t!]
  \centering
  
  \begin{subfigure}{\linewidth}
    \includegraphics[width=\textwidth, height=4.3cm]{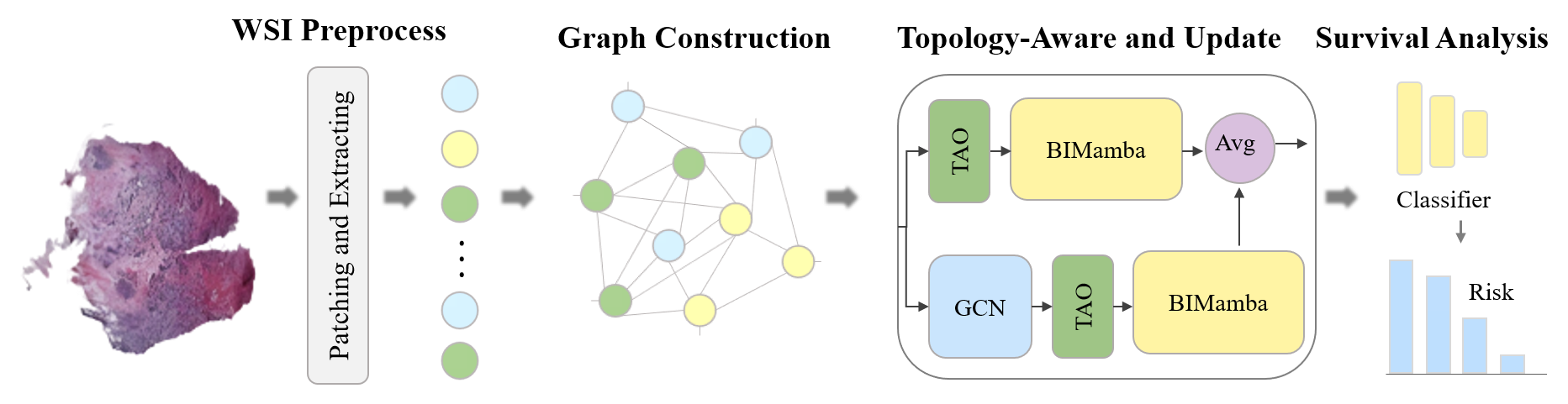} 
    \caption{The overview of TopoMamSurv framework}
    \label{fig:(a)}
  \end{subfigure}
  \vspace{10pt}

  \begin{subfigure}{0.49\linewidth}
    \includegraphics[width=\linewidth, height=4.5cm]{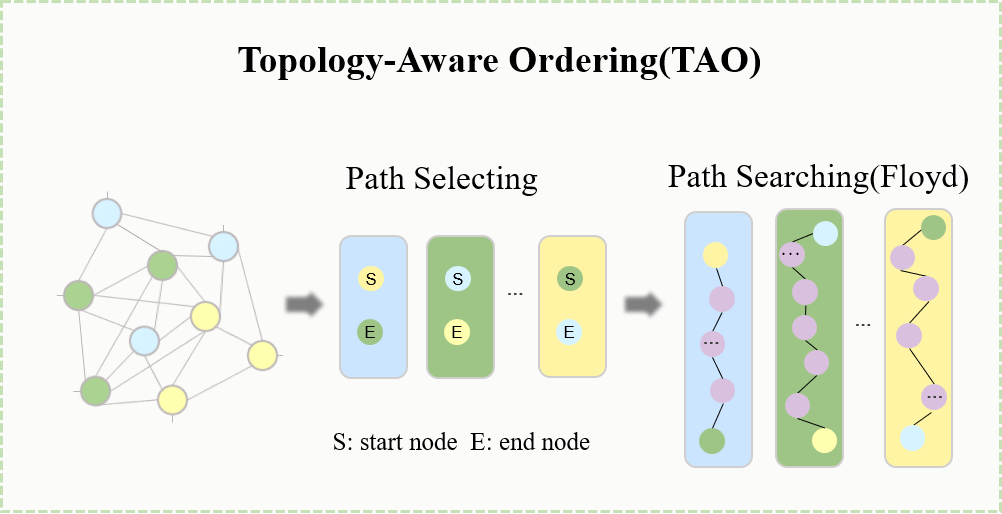} 
    \caption{Nodes Sort}
    \label{fig:(b)}
  \end{subfigure}\hfill
  \begin{subfigure}{0.49\linewidth}
    \includegraphics[width=\linewidth, height=4.5cm]{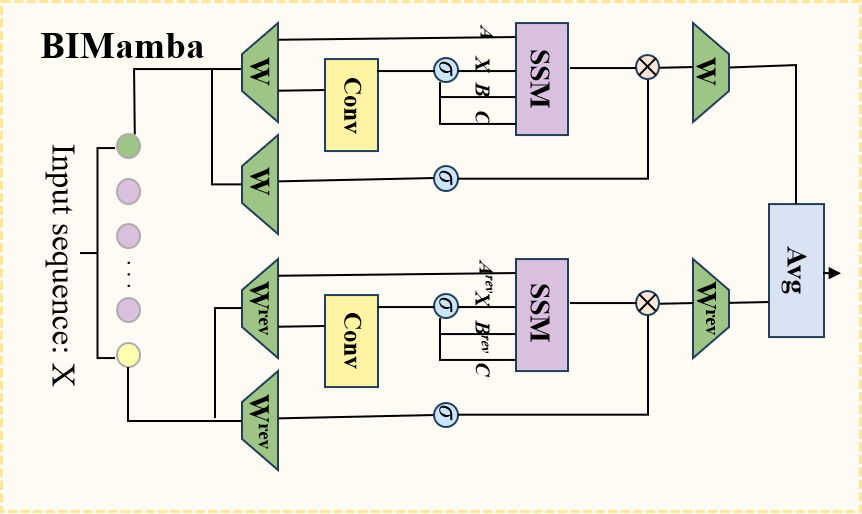} 
    \caption{Bidirectional-Mamba Module}
    \label{fig:(c)}
  \end{subfigure}

  \caption{\textbf{The TopoMamSurv framework.} \textbf{(a)} The overview of TopoMamSurv framework. First, features are extracted from WSI and optimized using a prototype attention mechanism. Subsequently, a fully connected graph is constructed and TAO is performed. The sorted nodes are then analyzed through a joint bidirectional Mamba-GCN module to generate survival predictions.
  \textbf{(b)} The method constructs path length matrices \(D_{ij}\) (adj) using edge weights, randomly selects multiple node pairs (e.g. A-E) as endpoints, dynamically updates shortest paths through recursive formulas \(D_{ij}^{(k)} = \min(D_{ij}^{(k-1)}, D_{ik}^{(k-1)} + D_{kj}^{(k-1)})\), and outputs topologically sorted node sequences (e.g. A-D-E-B).
  \textbf{(c)} The input X undergoes processing through convolutional kernel W and enters a standard Mamba layer to capture forward context, while being simultaneously fed into transposed components$(W^{\text{rev}}, A^T, B^T)$ for reverse state-space model computation. The dual-branch outputs undergo weighted aggregation at the gating layer before final representation generation through the linear layer.
  }
  \label{fig:main figure}
\end{figure*}

\textbf{MIL-based Methods.} MIL-based methods treat the patches from a WSI as multiple instances of one example. Traditional MIL-based methods, whether based on pooling~\cite{foncubierta2013medical,zhang2014towards} or bag-of-visual-words (bovw)~\cite{yu2017proceedings,lyu2019advances}, are predefined and untrainable. Inspired by the advances in deep representation learning, recent methods have combined them with trainable neural networks to get more discriminative features. Moreover, to capture the contribution of each instance to the overall representation, many studies have introduced attention mechanisms into MIL~\cite{ilse2018attention} and extended them to, for example, dual-stream attention~\cite{li2021dual} and clustering attention~\cite{xie2020beyond,guan2022node}. In addition, the Transformer model, with its specific self-attention mechanism, has been applied to effectively capture long-range dependencies and global features for MIL-based methods~\cite{shao2021transmil,lv2022transsurv,shao2023hvtsurv}. More recently, the newly developed state-space model, Mamba, has also been considered~\cite{yang2024mambamil,fang2024mammil}.

\textbf{Graph-based Methods.} Graph-based methods model the WSI as graphs, where one WSI is treated as one graph with patches or corresponding features as its nodes. Based on such a graph representation, message-passing GNN are mostly used to capture the spatial connectivity between different tissue regions. While early graph-based studies rely on handcrafted features~\cite{wang2021hierarchical}, pretrained deep Convolutional Neural Networks (CNNs) have become more prevalent as feature extractors~\cite{ding2022spatially,chen2021whole,sun2023tgmil}. Considering the advantages of Transformer in long sequence modeling, many studies have attempted to combine GNNs with Transformer to mitigate the over-smoothing issue in GNNs and to better capture global node connectivity~\cite{sun2023tgmil,zheng2022graph}. The recent emergence of the Mamba model, with its notable advantages, has also attracted researchers to integrate graph data with Mamba for further investigation~\cite{ding2025combining,yang2024graphmamba}.

\subsection{Long Sequence Modeling}
Long-sequence modeling is crucial in many fields, such as natural language processing and time-series forecasting~\cite{liu2014recursive,patro2024simba,zhang2023survey}. To this aim, Transformer~\cite{vaswani2017attention} has been designed and shown its effectiveness in wide applications ~\cite{nie2022time,zhou2023beyond,liu2021swin,dosovitskiy2020image}. However, such Transformer-based models have heavy computational and memory burdens and low efficiency in long-sequence modeling due to their \(O(N^2)\) time complexity~\cite{zhou2021informer,liu2022pyraformer}. Therefore, some studies have attempted to reduce computational complexity and memory consumption by modifying the way attention is calculated, such as sparse attention~\cite{zhou2021informer} and pyramid attention~\cite{liu2022pyraformer}. However, experience has shown that these approximate methods may perform poorly on large-scale sequences as they weaken the key advantages of the Transformer~\cite{gu2023mamba}. To overcome these limitations, the Mamba model has been proposed, which is based on a selective state-space model~\cite{gu2023mamba}. By parameterizing the input, it can filter information and retain key data. Combined with hardware-aware algorithm optimization, Mamba achieves linear scalability and the modeling power of the Transformer, showing excellent performance in both language~\cite{le2024locost,lieber2024jamba} and vision~\cite{liu2024vmamba,ma2024u,zhu2024vision} tasks. Inspired by its ability to model long-range dependencies, Mamba has also been applied to the WSI-based survival analysis task, with both MIL- and Graph-based methods, as reviewed in Section \ref{sec:WSI_SA}.

\subsection{Ordering Methods for Graph Mamba}
When dealing with graph-based representation with Mamba~\cite{li2024stg,pan2024hetegraph,wang2024graph,behrouz2024graph}, it is crucial to design a proper way of ordering when flattening the graph into sequences. Currently, several strategies have been considered. For example, Wang et al.~\cite{wang2024graph} have proposed ordering the nodes according to their degrees, and Behrouz et al.~\cite{behrouz2024graph} have attempted to order the sequences with the sizes of subgraphs. However, such methods ignore the topological structures of the graph. For example, the neighboring nodes would not necessarily be neighbors in the flattened sequence if we order the nodes according to their degrees. Consequently, the local relationship between nodes would be broken. Therefore, a proper sorting strategy for the graph Mamba still needs exploration.

\section{Problem Formulation}
\subsection{WSI-based Survival Analysis}
Suppose $P = \{p_1, p_2, \ldots, p_N\}$ is the WSI-based survival data of $N$ patients, where $p_i$ is a triplet $p_i = \{x_i, c_i, y_i\}$, with $x_i$ being the WSI, $c_i \in \{0, 1\}$ denoting the right censorship status ($c_i = 1$ means that the $i$-th sample is uncensored and $c_i = 0$ otherwise), and $y_i$ recording the observed time. The goal is to estimate the following hazard function $h(y|x)$:
\begin{equation}
h(y|x) = \lim_{\Delta y \to 0} \frac{P(y \leq O \leq y + \Delta y | O \geq y, x)}{\Delta y},
\end{equation}
where $O$ is the survival time, and consequently obtain the survival function $S(y|x) = P(O \geq y|x)$.

In deep learning practice for WSI-based survival analysis, the observed time is often discretized into intervals, such that $y_i$ takes the integer value indicating which time interval (or the risk level) the $i$-th patient belongs to ~\cite{zhao2023survival,zhou2023cross}. Consequently, the hazard function becomes
\begin{equation}
h(y|x) = P(O = y | O \geq y, x), \quad y = 1, 2, \cdots, K,
\end{equation}
where $K$ denotes the total risk levels. Then the survival function can be obtained as
\begin{equation}
S(y|x) = \prod\nolimits_{j=1}^{y} (1 - h(j|x)).
\end{equation}

With the above formulation, $h$, which is often parameterized as a deep neural network~\cite{shou2024graph}, can then be learned by minimizing the following loss derived from the maximum likelihood estimation:
\begin{equation}
\begin{split}
\mathcal{L}_{\text{surv}} = & -\sum\nolimits_{i=1}^{N} c_i \left[ \log S(y_i|x_i) + \log h(y_i|x_i) \right] \\
& - \sum\nolimits_{i=1}^{N} (1 - c_i) \log S(y_i + 1|x_i)
\end{split}.
\end{equation}
\subsection{Graph Representation for WSI-based Survival Analysis}\label{sec:graph_WSISurv}
As mentioned before, graph-based methods~\cite{ding2022spatially,chen2021whole,sun2023tgmil,wang2021hierarchical} have achieved promising performance for WSI-based survival analysis, due to the ability of the graph representation in capturing complex structures and contextual information to reflect pathological features and their interaction, which results in good global representations for WSIs. The main idea is to first divide the large WSI into a series of non-overlapping patches and extract features for each of them, and then construct a graph where each node represents the feature of one patch. Each edge represents the neighborhood relationship between two patches within the graph. Such a WSI graph can be formally defined as $G = (V, E, F)$, where $V$ denotes the node set, $E$ denotes the edge set, and $F \in \mathbb{R}^{|V| \times d}$ is the node attribute matrix, with $|V|$ being the total number of nodes, and $d$ the dimension of the attribute (or feature) of each node. With the constructed WSI graphs, the original WSI-based survival analysis can now be regarded as a graph-based learning problem using data triplets $\{G_i, c_i, y_i\}_{i=1}^N$, where $G_i$ is the graph constructed by WSI $x_i$. In this work, we adopt CLAM \cite{lu2021data} to crop each WSI into non-overlapping patches, and then utilize a pre-trained ResNet-50 model to extract 1024-dimensional features from each patch~\cite{chen2021multimodal,zhou2023cross,lu2021ai}. To reduce the complexity while retaining key information, we employ a prototype-guided attention mechanism~\cite{arik2020protoattend} to select the most representative nodes (prototypes). We then finally represent each WSI as a fully connected weighted graph with the selected nodes.

\section{Method}
\subsection{State-Space Model and Mamba}

\begin{table*}[t]
\centering 
\setlength{\tabcolsep}{10pt} 
\begin{tabular}{@{}lccccc@{}}
\toprule
\textbf{Methods} & \textbf{BLCA} & \textbf{BRCA} & \textbf{GBMLGG} & \textbf{LUAD} & \textbf{UCEC} \\ 
\midrule
DeepGraphConv & 0.580 ± 0.043 & 0.550 ± 0.099 & 0.785 ± 0.008 & 0.592 ± 0.045 & 0.631 ± 0.065 \\
Patch-GCN & 0.562 ± 0.042 & 0.600 ± 0.035 & 0.806 ± 0.016 & 0.598 ± 0.053 & 0.646 ± 0.047 \\
HGT & 0.575 ± 0.092 & 0.639 ± 0.070 & 0.853 ± 0.039 & 0.642 ± 0.049 & 0.661 ± 0.052 \\
CLAM-MB & 0.512 ± 0.029 & 0.580 ± 0.087 & 0.736 ± 0.052 & 0.567 ± 0.038 & 0.571 ± 0.051 \\
CLAM-SB & 0.528 ± 0.027 & 0.546 ± 0.070 & 0.745 ± 0.036 & 0.574 ± 0.050 & 0.586 ± 0.080 \\
TransMIL & 0.561 ± 0.049 & 0.641 ± 0.079 & 0.852 ± 0.020 & 0.645 ± 0.069 & \underline{0.669 ± 0.110} \\
Patch-GCN+VarPool & 0.571 ± 0.023 & 0.607 ± 0.069 & 0.814 ± 0.021 & 0.605 ± 0.045 & 0.655 ± 0.040 \\
GraphLSurv & 0.551 ± 0.013 & 0.586 ± 0.052 & 0.787 ± 0.021 & 0.585 ± 0.064 & 0.649 ± 0.043 \\
GraphMamba & \underline{0.612 ± 0.025} & \underline{0.664 ± 0.008} & 0.807 ± 0.033 & \underline{0.647 ± 0.019} & 0.663 ± 0.018 \\
MambaMIL & 0.588 ± 0.039 & 0.581 ± 0.033 & \textbf{0.881 ± 0.024} & 0.567 ± 0.022 & 0.530 ± 0.084 \\

\midrule
\textbf{Ours} & \textbf{0.654 ± 0.016} & \textbf{0.686 ± 0.016} & \underline{0.863 ± 0.020} & \textbf{0.664 ± 0.011} & \textbf{0.701 ± 0.011} \\ 
\bottomrule
\end{tabular}

\captionsetup{justification=raggedright, singlelinecheck=false}
\caption{The performance of different approaches on five public TCGA datasets in terms of c-index. The \textbf{best} and \underline{second best} results are highlighted.}
\label{tab:main experiments}
\end{table*}

A standard form of State-Space Model (SSM) is as follows~\cite{huang2024av}:
\begin{equation}
\begin{aligned}
\dot{\mathit{h}_t} &= \mathit{A}\mathit{h}_t + \mathit{B}\mathit{x}_t, \\
\mathit{y}_t &= \mathit{C}\mathit{h}_t + \mathit{D}\mathit{x}_t,
\end{aligned}
\end{equation}
where $A \in \mathbb{R}^{N \times N}$, $B \in \mathbb{R}^{N \times 1}$, $C \in \mathbb{R}^{1 \times N}$, and $N$ is the dimension of the hidden states. The input function or sequence $x_t \in \mathbb{R}$ is mapped to the output function or sequence $y_t \in \mathbb{R}$ through the hidden states $h_t \in \mathbb{R}^N$. After applying zero-order hold discretization, the above equations can be transformed to
\begin{equation}
\begin{aligned}
\mathit{h}_t &= \mathit{\overline{A}h}_{t-1} + \mathit{\overline{B}x}_t, \\
\mathit{y}_t &= \mathit{C h}_t + \mathit{D x}_t,\\
\mathit{\overline{A}} &= \mathit{\exp(\Delta A)}, \\
\mathit{\overline{B}} &= \mathit{(\Delta A)^{-1}(\exp(\Delta A) - I) \cdot \Delta B},
\end{aligned}
\label{eq:SSM}
\end{equation}
which results in the structured state space sequence (S4) model \cite{gu2021efficiently}. The S4 is characterized by four parameters $(\Delta, A, B, C)$ and possesses a significant property known as linear time invariance, indicating that the model's dynamics remain constant over time. Such a property not only enhances computational efficiency but also constrains the model's capacity for context-aware reasoning.

The Mamba model facilitates the context-aware reasoning capability of SSMs by dynamically adjusting its parameters based on input relevance~\cite{gu2023mamba}:
\begin{equation}
\begin{aligned}
\mathit{B} &= \mathit{s_B(x)}, \\
\mathit{C} &= \mathit{s_C(x)}, \\
\mathit{\Delta} &= \mathit{\tau_\Delta(\text{Parameter} + s_\Delta(x))}.
\end{aligned}
\label{eq: liner parameters}
\end{equation}
Here, $s_B(x)=\text{Linear}_B(x)$, $s_C(x)=\text{Linear}_C(x)$, $s_\Delta(x)=\text{Broadcast}_\Delta(\text{Linear}_\Delta(x))$ and $\tau_\Delta=\text{softplus}$. Due to its long-sequence modeling capability, Mamba has been successfully applied to various tasks, such as recommendation systems \cite{su2024mlsa4rec,liu2024mamba4rec}, survival analysis \cite{zhang2024samamba,cui2026mgcm}, among others.

\subsection{Ordering in Graph Mamba}
Mamba was originally proposed to handle sequence data and can only take sequences as inputs. Therefore, when applying Mamba to other types of data, one needs first to reform the data into sequences. Consequently, the scanning order becomes a crucial issue when implementing Mamba~\cite{liu2025vision}. For image data, several scanning strategies have been proposed based on the grid structure of image patches, including bidirectional scan~\cite{zhu2024vision}, cross scan~\cite{liu2024vmamba}, and so on. However, when dealing with graph data, it turns out to be more difficult to specify the scanning order due to the more complex topological structure of graphs. Currently, there are mainly two approaches that have been attempted, i.e., degree-based ordering~\cite{fang2024mammil} and subgraph size-based ordering~\cite{yang2024graphmamba}, which are depicted in \cref{fig:Three order}. Specifically, the degree-based method orders the nodes according to their degrees to form the input sequences, while the subgraph size-based one first aggregates the node features within sampled subgraphs with random walks, and then orders the aggregated features according to the lengths of the walks. On the one hand, these methods can effectively capture the global information of the graphs to some extent due to the use of Mamba. On the other hand, they completely ignore the local topological relationships among graph nodes. For example, the neighboring nodes in the original graph could not necessarily be adjacent ones in the flattened sequences. Such an ignorance of the local topological structure could lead to the loss of crucial local correlations of the graph-represented features. Therefore, a more appropriate scanning order is worthy of investigation for graph Mamba. 

\subsection{Topology-Aware Ordering}
\label{Topology-Aware Scaning}
Our solution to the aforementioned scanning order issue in graph Mamba is to consider the local topology structures and generate sequences using the shortest paths between pairs of nodes. Specifically, for each pair of nodes, one for start and one for end, we can find a shortest path between them using well-developed algorithms, such as the Floyd algorithm~\cite{floyd1962algorithm}. Then we can construct the input sequence with all the nodes along the path, in the order from the start point to the end. Such a scanning order of nodes can have two advantages: on the one hand, since the adjacent nodes along the sequence are neighbors in the original graph, the local dependencies between nodes are expected to be retained; on the other hand, the non-local correlations between long-distance nodes are also likely to be captured through the shortest path. Consequently, both the local and global information of the graph can be expected to be effectively extracted and processed.

\begin{figure*}
    \centering
    \begin{subfigure}[b]{0.19\textwidth}
        \includegraphics[width=\linewidth]{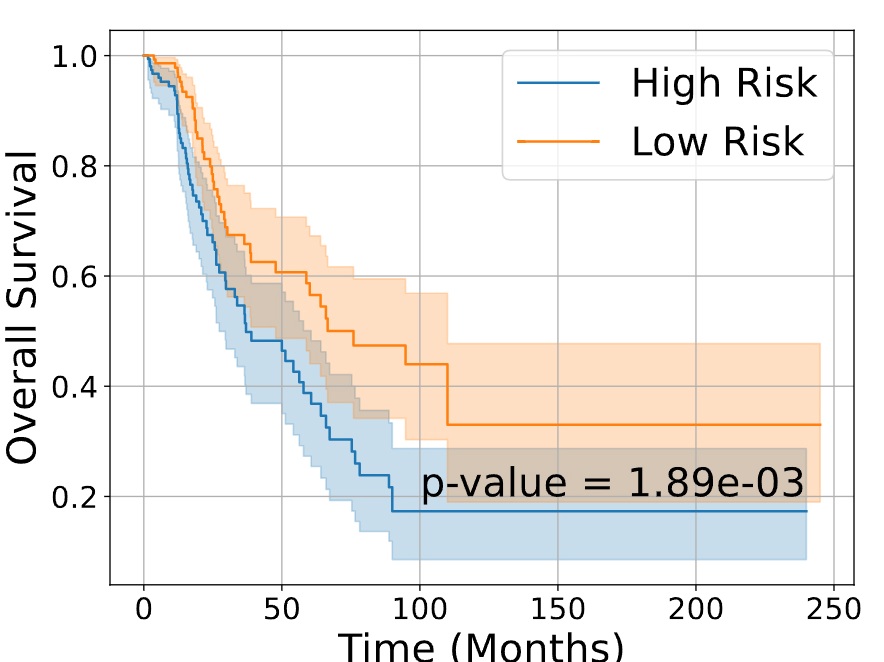}
        \caption{BLCA}
        \label{fig:blca}
    \end{subfigure}
    \hfill
    \begin{subfigure}[b]{0.19\textwidth}
        \includegraphics[width=\linewidth]{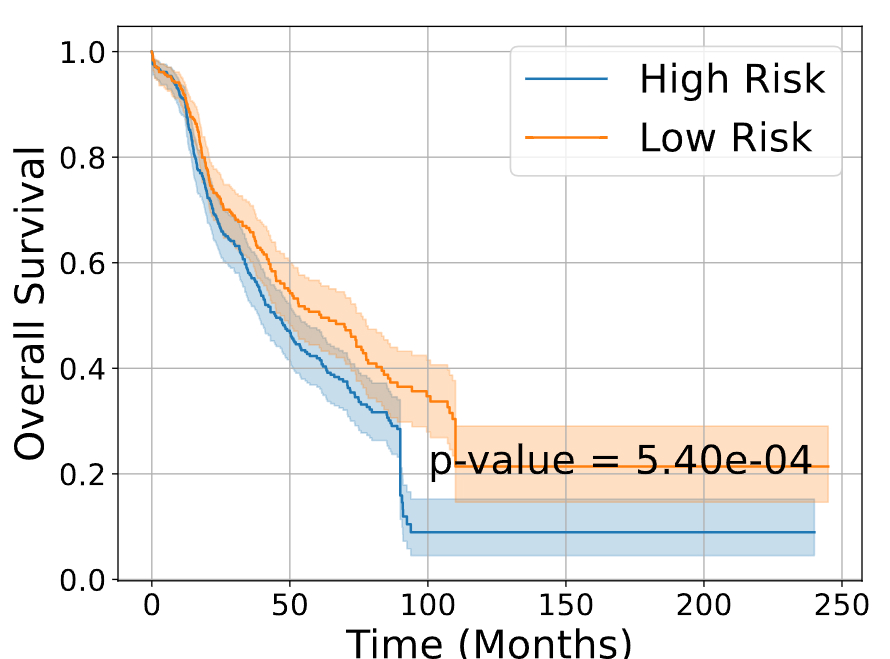}
        \caption{BRCA}
        \label{fig:brca}
    \end{subfigure}
    \hfill
    \begin{subfigure}[b]{0.19\textwidth}
        \includegraphics[width=\linewidth]{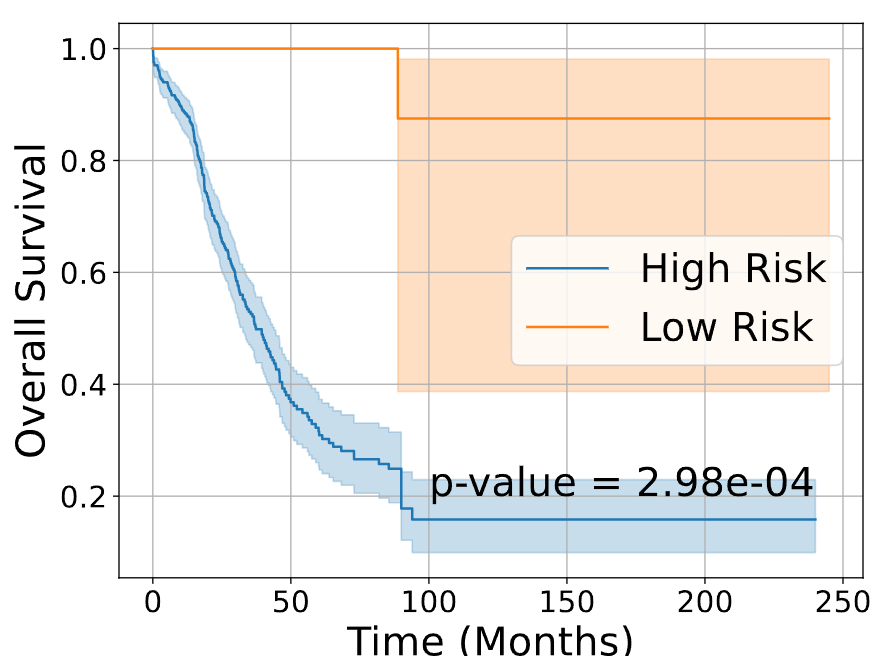}
        \caption{GBMLGG}
        \label{fig:gbmlgg}
    \end{subfigure}
    \hfill
    \begin{subfigure}[b]{0.19\textwidth}
        \includegraphics[width=\linewidth]{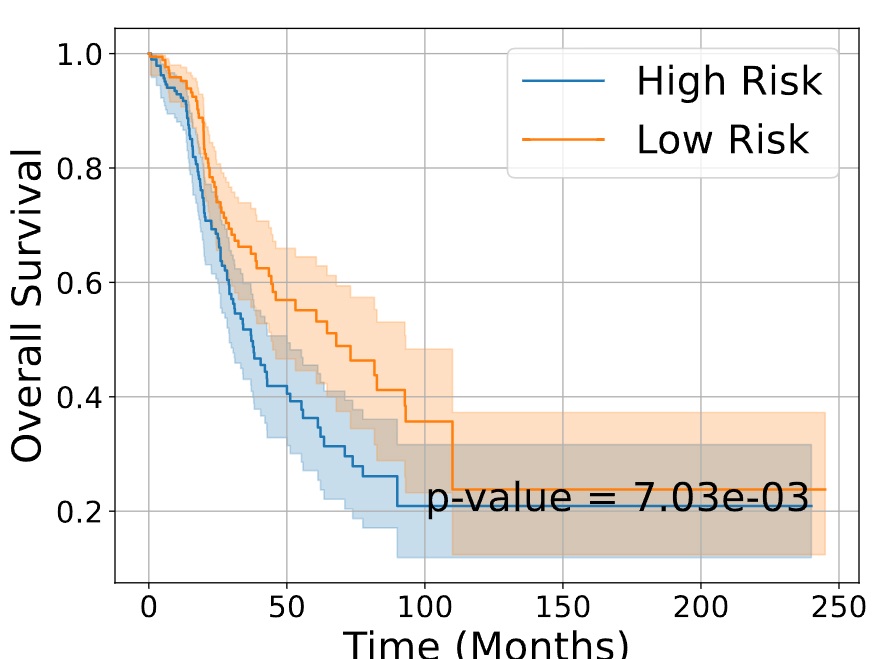}
        \caption{LUAD}
        \label{fig:luad}
    \end{subfigure}
    \hfill
    \begin{subfigure}[b]{0.19\textwidth}
        \includegraphics[width=\linewidth]{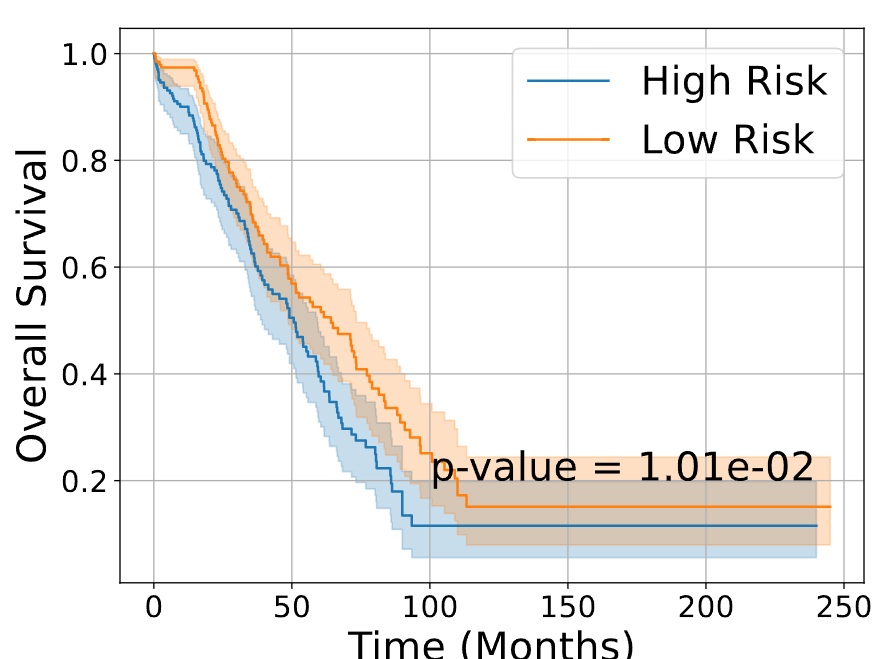}
        \caption{UCEC}
        \label{fig:ucec}
    \end{subfigure}

    \caption{
        Based on computed risk scores, patients are categorized into two groups: low-risk (orange) and high-risk (blue). The survival differences between these groups are then statistically evaluated using Kaplan-Meier estimation and Logrank testing.
    }
    \label{fig:survival_analysis}
\end{figure*}

Ideally, we can generate sequences for all pairs of nodes as inputs to Mamba. However, since the graph could be extremely large, such a way may result in unaffordable computational cost. Therefore, in practice, we can randomly sample a fixed number of pairs. We have designed a specific node-pairing strategy that ensures the tail node of each node pair is identical to the head node of the next node pair. By employing a multi-segment path concatenation approach, we construct coherent long-path structures while avoiding redundant node sampling. We empirically find that this random sampling strategy has a promising balance between efficiency and performance. The overall topology-aware ordering process for graph Mamba is illustrated in \cref{fig:Three order}. 

\subsection{Network Design}
\label{Dual-Branch feature update}
Using the graph Mamba with the proposed TAO method as the main building block, we now present our network design for WSI-based survival prediction, whose workflow is illustrated in \cref{fig:(a)}. Specifically, the WSI-oriented graph is first constructed as discussed in Section \ref{sec:graph_WSISurv}. Then, a Topology-Aware and Update module, which incorporates the proposed topology-aware sorting strategy with Mamba, is designed for processing and aggregating the graph-based features. Finally, a classifier is used for survival analysis using the resulting features. For the Topology-Aware and Update module, we employ a multi-branch and bi-directional structure for better processing of the features:

\textbf{Multi-branch structure.} In addition to directly inputting the extracted sequences, which are obtained using the TAO method, to the Mamba block, we consider extra branches that first process the original WSI graph with GCN blocks, then applies the TAO method again to extract new sequences from the processed graph, and finally inputs these sequences into another Mamba block. Such a design aims for a higher-order interaction among nodes. In our experiments, we empirically find that a two-branch structure, with one GCN block for the extra branch, gives the best performance.

\textbf{Bi-diretional Mamba.} Since we model the WSIs as unidirectional graphs, the order of the sequence can be naturally reversed. Therefore, we adopt the bi-directional technique~\cite{zhu2024vision} for a more effective feature update.

\section{Experiments}
In this section, we present experimental results to show the effectiveness of the proposed method. We first introduce the datasets and evaluation metrics, as well as some implementation details of our method. Then, we show the quantitative results on the adopted benchmarks. After that, we present further analysis to reveal the reasonability of the design mechanism for our method.

\subsection{Experimental Settings}

\subsubsection{Datasets}
Launched in 2006 by the National Cancer Institute and the National Human Genome Research Institute, The Cancer Genome Atlas (TCGA) project covers 33 types of cancer, analyzes over 11,000 samples, and provides data on clinical information, genomics, transcriptomics, and proteomics, thereby facilitating cancer research and precision medicine. In this study, we selected five cancer datasets from TCGA: bladder urothelial carcinoma (BLCA, $n=373$), breast invasive carcinoma (BRCA, $n=956$), glioblastoma multiforme and lower grade glioma (GBMLGG, $n=569$), lung adenocarcinoma (LUAD, $n=452$), and uterine corpus endometrial carcinoma (UCEC, $n=480$), to test prognostic models. We used five-fold cross-validation to assess model performance and compared it with other models.

\subsubsection{Evaluation Metrics}
Concordance Index (C-index) is a metric used to evaluate the predictive ability of a model. It is mainly used in survival analysis, measuring the model's ability to correctly order pairs of individuals based on predicted survival times. It can be represented by the following formula:
\begin{equation}
c\text{-index} = \frac{1}{n(n-1)} \sum_{i=1}^{n}\sum_{j=1}^{n} I(T_i < T_j)(1-c_j)
\end{equation}
Here, \( n \) represents the number of samples. \( T_i \) and \( T_j \) denote the survival times of the \( i \)-th and \( j \)-th samples, respectively.The function \( I(\cdot) \) is an indicator function that takes the value of 1 if the argument is true, and 0 otherwise.\( c_j \) indicates the right-censoring status of the \( j \)-th sample.


\subsubsection{Competing Methods}
For a comprehensive comparison, we adopt 10 existing methods in our experiments, including two Mamba-based methods: GraphMamba~\cite{yang2024graphmamba} and MambaMIL~\cite{yang2024mambamil}, and 8 representative non-Mamba deep learning methods: TransMIL~\cite{shao2021transmil}, CLAM\_MB~\cite{lu2021data}, CLAM\_SB~\cite{lu2021data}, Patch-GCN~\cite{chen2021whole}, PatchGCN+VarPool~\cite{carmichael2022incorporating}, GraphLSurV~\cite{liu2023graphlsurv}, DeepGraphConv~\cite{li2018graph} and HGT~\cite{hou2023multi}. Our method is referred to as TopoMamSurv.

\subsection{Results}
\begin{table*}[t]
\centering 
\setlength{\tabcolsep}{10pt} 
\begin{tabular}{@{}lccccc@{}}
\toprule
\textbf{Methods} & \textbf{BLCA} & \textbf{BRCA} & \textbf{GBMLGG} & \textbf{LUAD} & \textbf{UCEC} \\ 
\midrule
degree & 0.636 ± 0.024 & 0.679 ± 0.019 & 0.837 ± 0.016 & 0.641 ± 0.060 & 0.675 ± 0.054 \\
subgraph & 0.634 ± 0.035 & 0.675 ± 0.020 & 0.839 ± 0.012 & 0.636 ± 0.055 & 0.672 ± 0.023 \\
shuffle & 0.615 ± 0.022 & 0.662 ± 0.043 & 0.825 ± 0.017 & 0.637 ± 0.014 & 0.682 ± 0.046 \\
\midrule
w/o Mamba branch & 0.648 ± 0.030 & 0.669 ± 0.052 & 0.836 ± 0.012 & 0.626 ± 0.017 & 0.676 ± 0.060 \\
w/o GCN-Mam branch & 0.616 ± 0.035 & 0.653 ± 0.029 & 0.824 ± 0.013 & 0.621 ± 0.038 & 0.672 ± 0.038 \\
w/o BiMam & 0.589 ± 0.032 & 0.656 ± 0.037 & 0.823 ± 0.020 & 0.618 ± 0.026 & 0.637 ± 0.023 \\
\midrule
GCN2-Mam & 0.613 ± 0.019 & 0.628 ± 0.018 & 0.826 ± 0.022 & 0.611 ± 0.020 & 0.682 ± 0.020 \\
\midrule
\textbf{Ours} & \textbf{0.654 ± 0.016} & \textbf{0.686 ± 0.016} & \textbf{0.863 ± 0.020} & \textbf{0.664 ± 0.011} & \textbf{0.701 ± 0.011} \\ 
\bottomrule
\end{tabular}

\captionsetup{justification=raggedright, singlelinecheck=false}
\caption{Ablation experiment results of different sorting strategies and modules in our method.}
\label{tab:sort and modules}
\end{table*}

\subsubsection{Comparisons with State-of-the-Art}




The performance of the proposed TopoMamSurv method,  against current state-of-the-art (SOTA)  WSI-based survival analysis methods across 5 TCGA datasets, is reported in Table \ref{tab:main experiments}.
As observed, our topology-aware graph Mamba-based approach consistently outperforms competing baselines, demonstrating strong generalizability across diverse cancer types. Specifically, TopoMamSurv achieves the highest C-index scores on four datasets: 65.4\% on BLCA, 68.6\% on BRCA, 70.1\% on UCEC, and 66.4\% on LUAD, surpassing existing Mamba- and transformer-based survival models in each case. On the challenging GBMLGG dataset, our method still attains a highly competitive C-index of 86.3\%, which remains on par with leading SOTA approaches. These consistent performance gains across multiple datasets confirm the effectiveness of our topology-aware ordering strategy for graph Mambda, and validate the robustness and clinical utility of the proposed framework.

\subsubsection{Ablation Studies}
\label{ablation}

In this section, we conduct supplementary experiments to further show the effectiveness of the designed TAO and modules by comparing it with several variants. 

\textbf{Impact of Sorting Methods.} In the proposed TAO, we use the shortest path method to sort nodes, which preserves the inherent sequential dependencies of the graph topology and enables Mamba to model both local and long-range structural patterns effectively. To investigate how alternative strategies affect performance, we evaluate three commonly adopted sorting variants: degree sort, which orders nodes by their connectivity but ignores global topological context; subgraph sort, which clusters nodes into local subgraphs and thus breaks long-range dependencies; and shuffle sort, which randomly permutes nodes and eliminates all structural order\cite{wang2024graph, behrouz2024graph}. As shown in Table \ref{tab:sort and modules}, all three alternatives lead to significant performance drops, confirming that the shortest path ordering is critical for our framework to capture comprehensive structural information and achieve optimal prediction.


\textbf{Impact of Modules.} In this part, we study the effectiveness of each core module by removing the pure Mamba branch, the GCN-Mamba branch, and the bidirectional mechanism of Mamba, respectively. The experimental results, as shown in Table \ref{tab:sort and modules}, demonstrate that discarding any of these components leads to clear performance degradation: removing the pure Mamba branch deprives the model of strong long-range sequential modeling ability, removing the GCN-Mamba branch weakens the integration of structural and sequential information, and disabling the bidirectional mechanism impairs the model’s ability to capture dependencies in both directions. These observations confirm that each module plays an indispensable role, and their synergy is key to the superior performance of our framework.

\textbf{Impact of Multi-Branch Ensemble.} To explore whether introducing more graph-enhanced branches can further boost performance, we design an extended three-branch variant by adding an extra branch with two consecutive GCN layers and one Mamba module. We then fuse predictions from all three branches for final inference. As shown in Table \ref{tab:sort and modules}, this three-branch variant (GCN2-Mam) yields lower performance than our original two-branch framework. This observation demonstrates that simply stacking additional graph-Mamba branches does not guarantee performance gains, and our concise two-branch design achieves a better trade-off between model complexity and efficiency.

\subsubsection{Further Analysis of Sorting Methods} To further analyze the reasonability of the proposed sorting strategy, we compute the similarity along the extracted node sequences using different sorting methods. Specifically, we have employed cosine similarity and normalized Euclidean distance as metrics for similarity measurement to facilitate comparison on the same scale. 
For each image, we randomly extract 30 sets of nodes from the four different sorting methods and calculate the average similarity among each set of nodes, using this as the node similarity result for that image. Finally, we averaged the similarity results across all 361 images in the LUAD dataset to derive the overall node similarity.



\begin{figure}
  \centering
  \includegraphics[width=\linewidth]{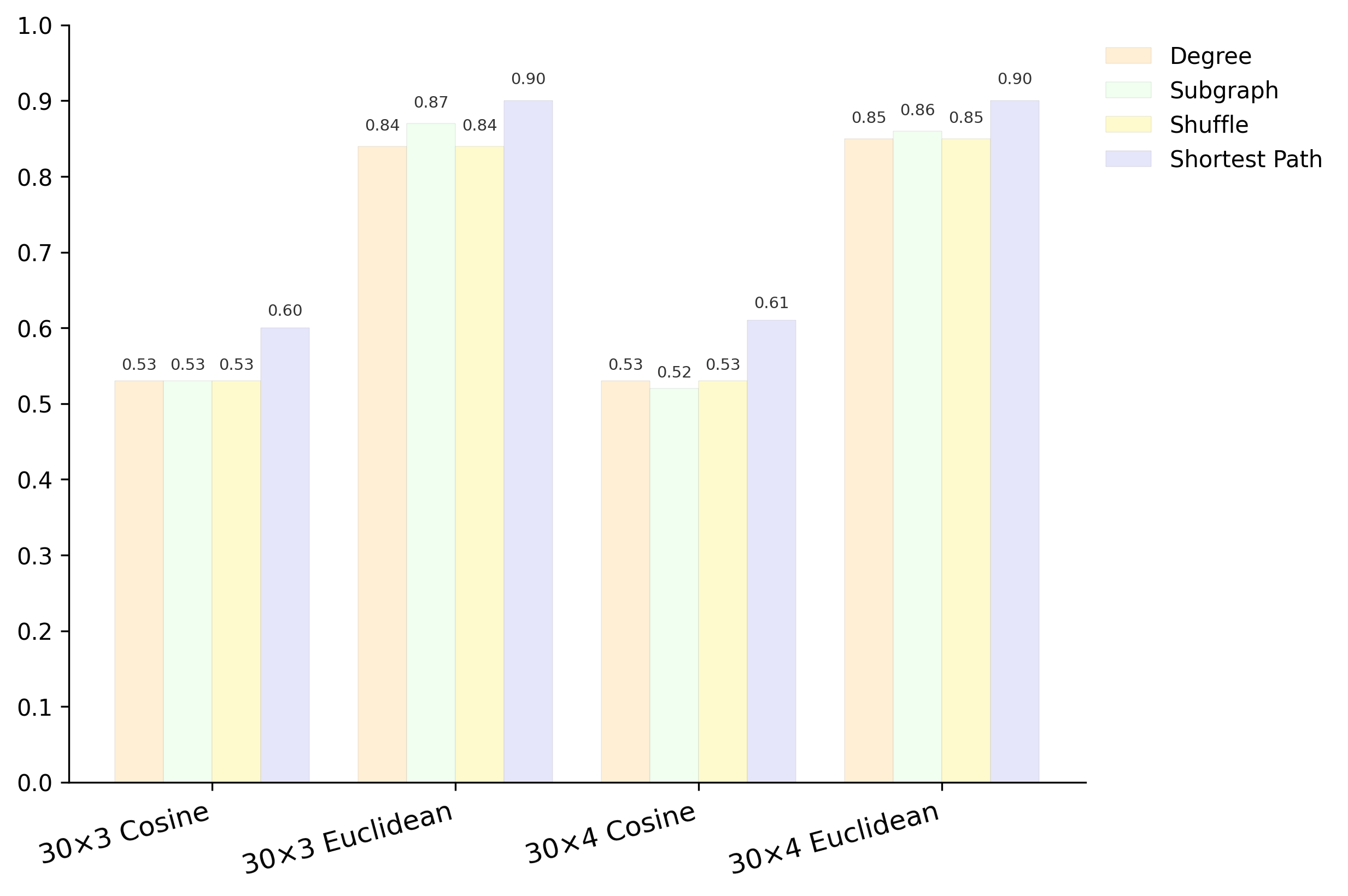}
  \caption{
    Performance of Four Sorting Strategies under Different Similarity Metrics 
  }
  \label{fig:similarity_results}
\end{figure}

As shown in Figure \ref{fig:similarity_results}, compared with degree sorting, subgraph sorting, and shuffle sorting, the node similarity selected by shortest path is higher than other methods. These results substantiate the efficacy of the proposed TAO strategy in capturing node similarity and underscore its potential to enhance the comprehension of node relationships in graph-based analyses, thereby offering significant insights for future research.

\subsubsection{Survival Analysis}
To assess the effectiveness of TopoMamSurv in survival analysis, we categorized patients into low-risk and high-risk groups based on the median of the predicted risk scores by TopoMamSurv. Subsequently, we employed the Kaplan-Meier analysis method to visually present the survival events of all patients, with the results depicted in Figure \ref{fig:survival_analysis}. Additionally, we used the Log-rank test ($p$-value) to evaluate the statistical significance of differences between the low-risk group (orange) and the high-risk group (blue). It is generally accepted that a $p$-value less than or equal to 0.05 indicates a statistically significant difference. From the figure, it is evident that our method yields $p$-values significantly below 0.05 across all datasets, confirming the validity of our results.

\section{Conclusion}
In this study, we propose a novel Graph Mamba survival analysis framework based on topology-aware ordering (TopoMamSurv) to adapt to the sequential sensitivity of Mamba, aiming to significantly enhance the performance of survival analysis for cancer patients based on histopathological images. Specifically, we employed a topology-aware approach, sorting nodes by identifying shortest paths within the graph structure. This method enables the sequential Mamba model to more comprehensively capture the complex relationships among numerous morphologically diverse patches within each whole slide image, thereby allowing for a more accurate understanding and prediction of patients' survival status. Experiments on five TCGA datasets have been conducted against existing methods, demonstrating the effectiveness of the proposed method. In the future, we plan to investigate more applications of the proposed sorting strategies for graph Mamba-based deep models

\bibliographystyle{IEEEtran}
\bibliography{myref}

\end{document}